\documentclass[letterpaper, 10 pt, conference]{ieeeconf}  

\IEEEoverridecommandlockouts                              

\usepackage{amsmath}
\usepackage{amssymb}
\usepackage{algorithm}
\usepackage{algpseudocode}
\usepackage[utf8x]{inputenc}
\usepackage{graphicx}

\title{\LARGE \bf
Visuo-Tactile Keypoint Correspondences for Object Manipulation
}

\author{Jeong-Jung Kim$^{1}$, Doo-Yeol Koh$^{1}$ and Chang-Hyun Kim$^{1}$
\thanks{* This study is a part of the research project, ``Development of core technologies for robot general purpose task artificial intelligence (RoGeTA) framework (NK248G)", which has been supported by a grant from National Research Council of Science \& Technology under the R\&D Program of Ministry of Science.}
\thanks{$^{1}$ The authors are with  Department of AI Machinery, Korea Institute of Machinery \& Materials, Daejeon, Korea
{\tt\small \{rightcore, dyk, chkim78\}@kimm.re.kr}}%
}

\begin{document}

\maketitle
\thispagestyle{empty}
\pagestyle{empty}

\begin{abstract}

    This paper presents a novel manipulation strategy that uses keypoint correspondences extracted from visuo-tactile sensor images to facilitate precise object manipulation. Our approach uses the visuo-tactile feedback to guide the robot's actions for accurate object grasping and placement, eliminating the need for post-grasp adjustments and extensive training. 
    This method provides an improvement in deployment efficiency, addressing the challenges of manipulation tasks in environments where object locations are not predefined.
    
    We validate the effectiveness of our strategy through experiments demonstrating the extraction of keypoint correspondences and their application to real-world tasks such as block alignment and gear insertion, which require millimeter-level precision. The results show an average error margin significantly lower than that of traditional vision-based methods, which is sufficient to achieve the target tasks.
\end{abstract}

\section{INTRODUCTION}

In the field of robotics, manipulation tasks that focus on the precise picking and placing of objects pose significant challenges, especially in environments where object locations are not predefined. Achieving precise manipulation in these environments requires advanced perception capabilities that allow robots to adapt their actions according to the identified object pose.

Traditionally, vision-based methods with color and depth cameras or LiDAR sensors have been used to estimate the pose of objects. However, these methods are often susceptible to sensor noise and environmental disturbances, which can affect the accuracy of manipulations. 
To overcome these limitations, research has explored the fusion of different sensing methods, including the novel application of visuo-tactile sensors. These sensors typically utilize a flexible elastomer material and a color sensor \cite{yuan2017gelsight, donlon2018gelslim, lambeta2020digit, taylor2022gelslim}. It enables the transformation of tactile data into visual images. These sensor components are attached to the robot's end effector or the tip of a gripper. This configuration allows for direct observation of the object's contact state during manipulation tasks. 

\begin{figure}[t]    \centering
    \includegraphics[scale=0.4]{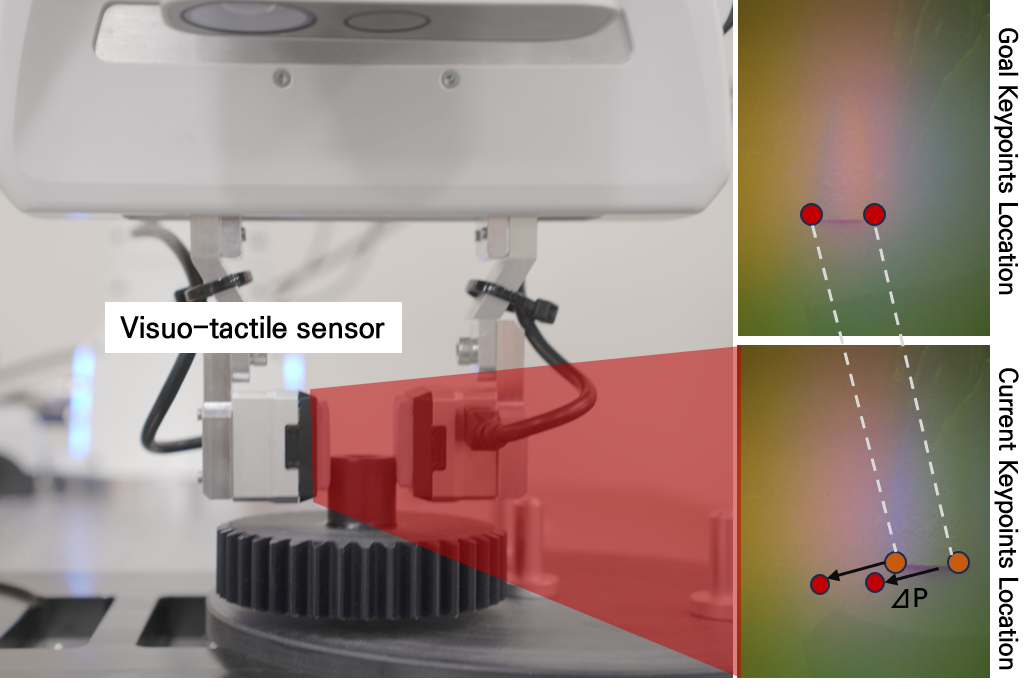}
    \caption{Displacement estimation based on keypoint correspondences from visuo-tactile sensor images for pose adjustment in robot manipulation}
    \label{figurelabel1}
 \end{figure}

Compared to traditional visual-based methods, such tactile sensors provide improved manipulation accuracy. The direct sensing of contact state through tactile feedback enhances the robot's ability to grasp and manipulate objects with greater precision. This visuo-tactile sensors have been applied to various tasks including grasping, part identification, pose refinement and stability assessment during manipulation, although challenges such as the need for extensive training \cite{dong2021tactile} and object marking \cite{lim2023grasping} for improved recognition remain.

This paper presents a novel manipulation approach that uses keypoint correspondences from images captured by a visuo-tactile sensor to guide manipulation. 
After extracting feature descriptors from both the goal image and the current acquired sensor image, we compare the values corresponding to predefined keypoints in the feature descriptors of the goal image with the similarities between the entire feature descriptors in the acquired image. Following this comparison, we proceed to select the point with the highest similarity for finding correspondences. We conduct displacement estimation based on the keypoint correspondences and pose adjustment for robot manipulation. This approach has two advantages: it eliminates the need for additional adjustments after grasping and eliminates the requirement for extensive training, making deployment more efficient and faster. 

The research has two contributions. Firstly, we propose a method that uses keypoint correspondences from visuo-tactile sensor data to enable precise manipulation without the need for additional learning. Secondly, we demonstrate the feasibility of this approach in real-world tasks, showing its effectiveness and reliability in enhancing manipulation precision.


The paper is organized as follows: Section II provides a detailed explanation of the proposed method, explaining the technical aspects and underlying principles. Section III presents the experimental setup and results, validating the efficacy of our approach across manipulation scenarios. Finally, Section IV concludes the paper by discussing the proposed approaches and their significance for manipulation, as well as suggesting future research directions.

\section{MANIPLUATION WITH TACTILE KEYPOINTS CORRESPONDENCES}

This paper presents a novel framework for manipulation using keypoints extracted from images captured by visuo-tactile sensors. Keypoint correspondences have previously been shown to be effective for object pose estimation and adaptable across variations within object categories \cite{florence2019self, manuelli2019kpam}. Our research expands on this approach by utilising it for visuo-tactile sensor data, allowing for accurate manipulation tasks through focused interaction with the object of interest.


Visuo-tactile sensors have the advantage of focusing on the object in contact, which facilitates accurate position estimation. It also features self-illuminating components, enabling superior performance even in changing lighting environments. Additionally, our method utilizes foundation models, which are pre-trained deep learning models capable of understanding a wide range of data patterns and features. These foundation models offer several advantages, including their ability to identify features without requiring object-specific training. This not only simplifies the process but also enhances the system's adaptability across diverse tasks and domains, thereby increasing its versatility and usability.

Thus, our framework's applicability to a wide range of objects and tasks is greatly enhanced by this aspect, without the need for extensive learning phases for each new object category.

\subsection{Overall Procedure}

\begin{figure*}[thpb]
    \centering
    \includegraphics[scale=0.4]{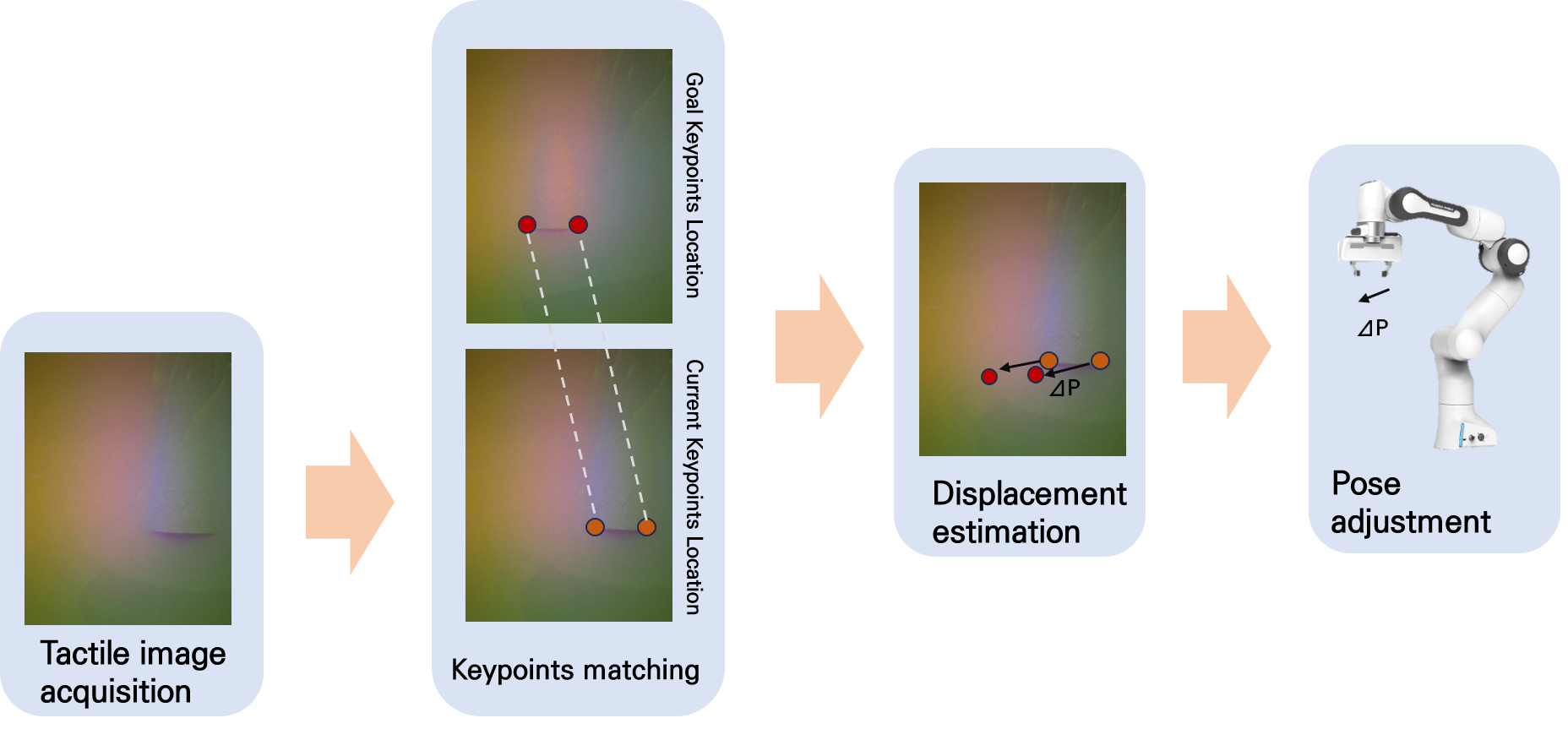}
    \caption{Manipulation process using keypoints extracted from visuo-tactile sensor images. The correspondence points between the sensor image obtained from human demonstration and the image captured during actual execution are identified. Displacement is calculated using this correspondence, and a pose adjustment is performed based on the value.}
    \label{overall_flow}
   \end{figure*}

This section describes a two-phase manipulation process that uses visuo-tactile sensing and keypoint correspondence to achieve precise object handling. Our approach assumes that when an object is grasped, its features, such as points, lines, and textures, can be observed. These features can then be aligned between two images to establish correspondences. 
The proposed method is illustrated in Fig. \ref{overall_flow} and detailed in Algorithm 1. 

\begin{algorithm}[t]
    \caption{Manipulation Algorithm with Visuo-Tactile Keypoints Correspondences}
    \begin{algorithmic}[1]
    \State \textbf{Initialize:} Threshold for displacement $\tau$
    
    \State \textbf{Demonstration Phase:}
    \State 1) Human demonstrator positions the gripper and captures $\textbf{I}_{g}$ stores gripper width $w$.
    \State 2) Define keypoints $\textbf{k}_{g} = \{(u_i, v_i)\}_{i=1}^{K}$.
    
    \State \textbf{Execution Phase:}
    \State 1) Attempt object grip and capture $\textbf{I}_{c}$.
    \State 2) Process images through dense descriptor model: \\
            \hspace*{1cm} $f^D(\textbf{I}_{g})$ and $f^D(\textbf{I}_{c})$.
    \State  3) Apply correspondence function for finding correspondences between $\textbf{k}_{g}$ and keypoints in $\textbf{I}_{c}$:\\ 
            \hspace*{1cm} $\textbf{k}_{c} = f^C(f^D(\textbf{I}_{g}), f^D(\textbf{I}_{c}), \textbf{k}_{g} )$.
    \State  4) Estimate displacement:\\
            \hspace*{1cm}    $\varDelta \textbf{P} = \text{EstimateDisplacement}(\textbf{k}_{g}, \textbf{k}_{c})$.

    \If{$|\varDelta \textbf{P}| < \tau$}
        \State \textbf{terminate} with success.
    \Else
        \State Adjust robot's end-effector pose in the Cartesian coordinate system by $\varDelta \textbf{P}$.
        \State Go to Step 1).
    \EndIf
    \end{algorithmic}
\end{algorithm}
    

    
    

Correspondence points are identified between the sensor image acquired from human demonstration and the image captured during actual execution. These points are then used to calculate displacement, enabling pose adjustment based on the obtained values.

\begin{itemize}

    \item Demonstration phase: 
In the first phase, a human demonstrates to the robot the grasp pose and grasping width required to perform a specific task. Although several approaches could be utilized for this purpose, our method involves the demonstrator manually positioning the robot's gripper to grasp the object. This action allows the visuo-tactile sensor to capture the object's shape, and keypoints are pre-defined and saved based on this captured image. In this process, we acquire a visuo-tactile image data $\textbf{I}_{g} \in \mathbb{R}^{W\times H\times C}$ containing the target pose suitable for a task, $K$ keypoints with $(u, v)$ pixel coordinates $ \textbf{k}_{g} = {\{u_{i}, v_{i}\}}^{K}_{i=1} $ designated by the demonstrator, and the gripper width $w\in \mathbb{R}$.

\item Execution phase: 
In the execution phase, the robot attempts to pick up the object using the approximate pose information obtained with a sensor such as a camera. It graps an object and then acquires a visuo-tactile image  $\textbf{I}_{c} \in \mathbb{R}^{W\times H\times C}$ from the sensor and identifies correspondences with the predefined keypoints $\textbf{k}_{g}$. 

Tactile images $\textbf{I}_{g}$ and $\textbf{I}_{c}$ are processed through a dense descriptor model $f^D(\cdot)$, followed by a correspondence function $f^C(\cdot)$, resulting in $\textbf{k}_{c}$, which corresponds to $\textbf{k}_{g}$.
Based on this correspondence, a displacement $\varDelta \textbf{P}$ is estimated, indicating the deviation from the target pose for manipulation in the Cartesian coordinate system.

If the norm of the displacement $|\varDelta \textbf{P}|$ is below a predefined threshold, the process terminates; otherwise, the robot releases the object and adjusts its position by the calculated displacement amount, repeating the process as necessary.

This iterative approach allows a more accurate determination of the grasp position and manipulation by taking into account how the object is held and adjusting accordingly. While a single attempt might fail due to recognition errors, iteration refines the robot's perception of the object's position and orientation.
This approach provides a refined strategy for manipulation, combining visuo-tactile feedback with visual feature matching to improve the accuracy and reliability of manipulation.

\end{itemize}

\subsection{Keypoints Correspondences}
To build dense descriptors for the tactile sensor data at the step 2 of the execution phase, we used the DINO, which uses a pre-trained Vision Transformer (ViT) to extract deep features that serve as dense visual descriptors \cite{caron2021emerging, amir2021deep}. These features capture strong, well-localised semantic information with a high degree of spatial granularity. Furthermore, the semantic information encoded in these features is applicable across a spectrum of related, yet distinct, object categories. In this paper, we used the DINO method to generate dense descriptors, but it is also possible to use other methods.

Depending on the characteristics of the object, more than one keypoint may be required. However, this paper focuses on using two keypoints to find correspondences, which is sufficient for calculating two-dimensional displacement in terms of position and angle. While three keypoints could allow for three-dimensional displacement calculations, the nature of visuo-tactile sensors limits the accuracy of depth measurements, making two-dimensional information more reliable for precise manipulation. However, since the number of keypoints required for pose estimation varies from object to object, it is necessary to adjust this parameter according to the specific problem at hand.

\begin{figure}[t]
    \centering
    \includegraphics[scale=0.15]{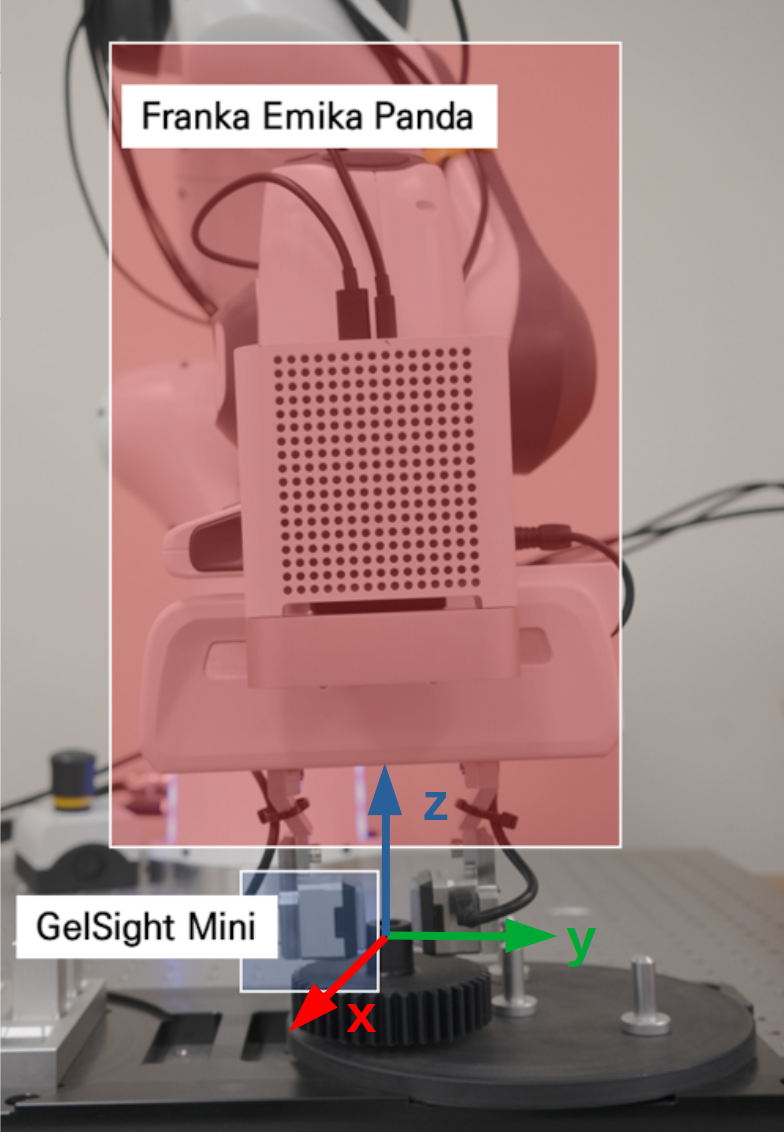}

    \caption{Experimental setup. A GelSight Mini sensor, which is a visuo-tactile sensor, is attached to the end-effector of the Franka Emika Panda robot to acquire sensor data and estimate displacement.}
    \label{setup}
 \end{figure}

 \begin{figure}[h]
    \centering
    \includegraphics[scale=0.16]{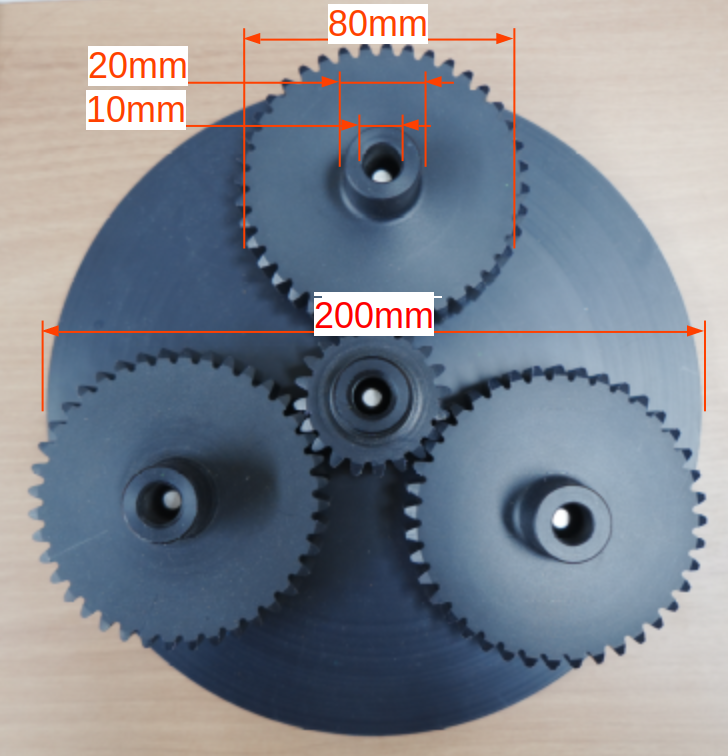}
    \caption{Objects for gear insertion task. A robot picks up gears and inserts them into holes on a panel.}
    \label{object}
 \end{figure}

    \section{Experiments}

    To investigate whether keypoint correspondences can be extracted from visuo-tactile images, the precision of the displacement estimation method, and whether this method can be applied to real manipulation tasks, we conducted a series of experiments.

Our experimental setup consisted of equipping a Franka Emika Panda from Franka Robotics \footnote{Franka Robotics, http://www.franka.de/} with a GelSight Mini sensor from GelSight \footnote{GelSight, http://www.gelsight.com/} at the gripper end of the robot, as shown in Fig. \ref{setup}. This sensor captures contact information within an area of $18mm \times 24mm$ at a resolution of $240 \times 320$, which we adjusted to $224\times 298$ for keypoint extraction. The experiment was conducted using only one of the two sensors attached to the robot. For feature extraction, we used the DINO method with the ViT-S/8 model, with a step size of 4.

The task performed by the robot and the size of the objects are shown in Fig. \ref{object}. This task involves picking up gears with holes and inserting them onto a shaft. The experiment was conducted with the gripper, equipped with sensors, grasping the upper part of the gear.
    
    \begin{figure}[t]
        \centering
        \includegraphics[scale=0.4]{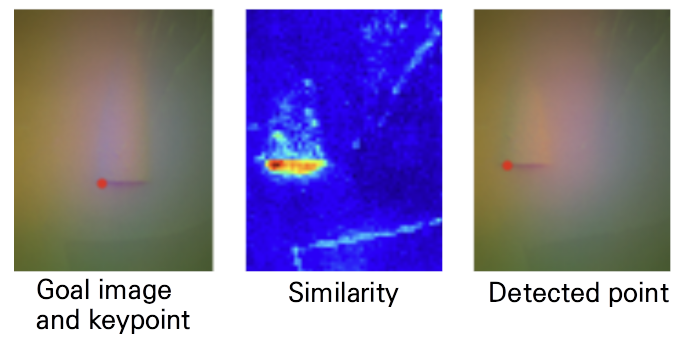}

        \caption{Example of successful keypoint correspondence. Keypoint matching has been performed, associating the left corner of the object in the goal image with the left corner of the object in a captured tactile sensor data.}
        \label{corr_ok}
     \end{figure}

     \begin{figure}[t]
        \centering
        \includegraphics[scale=0.4]{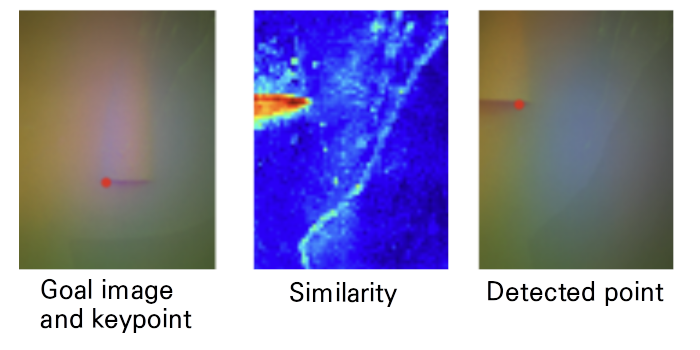}

        \caption{Example of unsuccessful keypoint correspondence. The keypoint matching has incorrectly associated the left corner of the object in the goal image with the right corner of the object in a captured tactile sensor data.}
        \label{corr_fail}
     \end{figure}

     \begin{figure*}[thpb]
        \centering
        \includegraphics[scale=0.8]{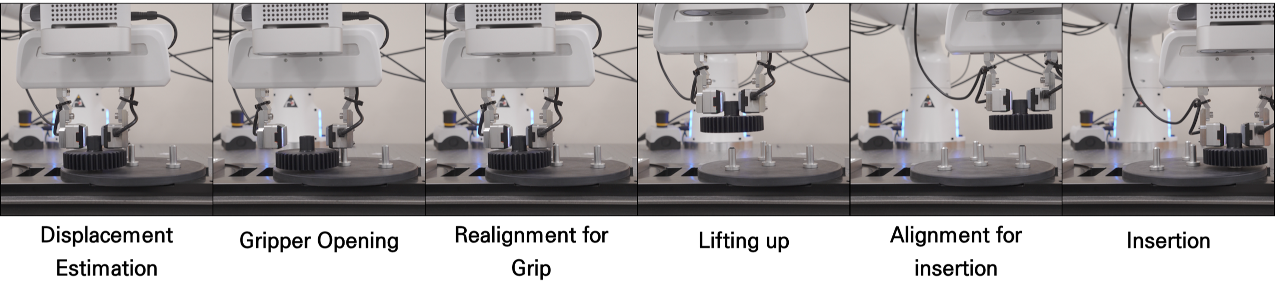}
        \caption{Snapshot of gear insertion task using the proposed method}
        \label{task1}
     \end{figure*}
     \begin{figure*}[thpb]
        \centering
        \includegraphics[scale=0.8]{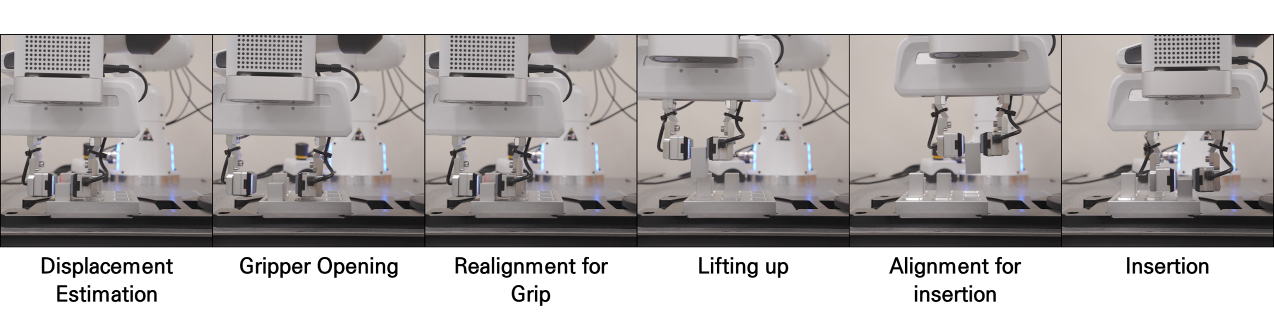}
        \caption{Snapshot of block alignment task using the proposed method}
        \label{task2}
     \end{figure*}

    \subsection{Keypoint correspondences}
    The first experiment aimed to verify the effectiveness of extracting keypoint correspondences from images captured by the visuo-tactile sensor, and to measure the deviation of these keypoints from their ground truth positions. The experiments are conducted targeting the task of grasping the gear at the appropriate position to ensure successful insertion. The robot's end-effector was manually moved and oriented to the pose where insertion should occur. At this pose, an image acquired from the tactile sensor were set as the goal image, and keypoints were manually defined on the goal image.
    
    We positioned the robot's end-effector at the pose and moved it randomly in a range of +/- 5mm in $x$ and $z$ axis to acquire test tactile images. The deviation between the keypoints extracted from these images and those identified by the operators was then evaluated. After extracting feature descritors using the DINO from both the goal image and the acquired sensor image, we compared the values corresponding to predefined keypoints in the descriptor of the goal image with the similarity between the entire descriptor in the acquired image. Following this comparison, we proceed to select the point with the highest similarity.
    

   The displacement estimation error was calculated with (\ref{error_eq}),
    \begin{equation}  \label{error_eq} 
        d_{error}=\frac{\sum_{i=1}^{N}\sqrt{ (p^{i}_{x_{real}} - p^{i}_{x_{est}})^2 + (p^{i}_{z_{real}} - p^{i}_{z_{est}})^2}}{N}
    \end{equation} 
    where $p^{i}_{x_{real}} $ and $p^{i}_{z_{real}}$ are real position of displacement which were measured from robot's end-effector position with kinematics information and $p^{i}_{x_{est}} $ and $p^{i}_{z_{est}}$ are estimated displacement calculated from suggested method, respectivley. 
    We conducted 10 experiments, and the average error was 1.29 mm and its standard deviation was 0.71mm.

    We found that in most cases the displacement estimation was succesed, as shown in Fig. \ref{corr_ok}. The left corner of the object in the goal image has been matched to the left corner of the object in the currently acquired image. However, when only part of the object was visible, or when it crossed the boundaries, keypoints corresponding to the opposite side were detected, leading to errors as shown in Fig. \ref{corr_fail}. The left corner of the object in the goal image has been incorrectly corresponded to the right corner of the object in the currently acquired image. One such case resulted in an error of 50 pixels, corresponding to an error of 3.75 mm. The limited detection range of this sensor leads to ambiguity at the boundary areas, resulting in such outcomes.    
    However, this error margin, which is relatively small compared to vision-based methods, is considered sufficient to achieve the target tasks using techniques such as impedance control, assuming a rough alignment between features such as lines and points in the captured and target images. 

     \subsection{Manipulation tasks}

To demonstrate the capability of the proposed method for precise tasks, we conducted experiments on gear insertion and block alignment tasks, both of which required millimeter-level accuracy that could not be achieved with external cameras alone. The method enabled alignment followed by robot control via impedance control. The experiments successfully confirmed the feasibility of both tasks, as shown in Figs. \ref{task1} and \ref{task2}. After gripping the object, the sensor image was acquired, and the displacement was estimated by performing keypoint correspondence.
From the obtained correspondences, the displacement is estimated. Based on this estimated displacement, an offset adjustment is applied to the robot's end-effector pose, enabling it to re-grasp the object and complete the task. In cases of insufficient pre-alignment, the use of sensors such as force-torque sensors and iterative search techniques or reinfocement learning are necessary to achieve correct positioning \cite{luo2019reinforcement}. However, our method reduces the burdens associated with completing the task, which is essential for applying the robot in real-world applications.

\section{CONCLUSION}
In this paper, we have introduced a manipulation strategy that uses keypoint correspondences from visuo-tactile sensor images to improve the precision of object picking and placement tasks. This method not only reduces the need for post-grasp adjustments, but also minimises the dependency on extensive training, thus increasing deployment efficiency.

The experimental results have validated the effectiveness of our approach, demonstrating that keypoint correspondences can be accurately extracted from visuo-tactile images, with an average positional error low enough to allow precise manipulation through techniques such as impedance control. Furthermore, our method has proven capable of performing tasks requiring millimeter-level accuracy, such as block alignment and gear insertion, which are challenging for traditional vision-based systems.

However, our method requires a rough initial alignment and the presence of detectable features for successful keypoint extraction. This could be a limitation in scenarios where such conditions are not met, suggesting the need for further research into active alignment strategies. In addition, the current approach requires predefined keypoints for each object category, which could be a drawback when dealing with new categories.


Future research will address these limitations through the development of algorithms that can actively adjust the position of the robot's end effector for optimal feature extraction. Additionally, automating the process of keypoint selection for new object categories would increase the versatility and applicability of our method, enabling robots to perform more complex and varied tasks in dynamic and unstructured environments.




\bibliographystyle{unsrt}
\bibliography{ref.bib}

\end{document}